\documentclass{article}

\usepackage{microtype}
\usepackage{graphicx}
\usepackage{booktabs} % for professional tables
\usepackage{amsmath,amssymb,mathtools,bm,etoolbox}
\usepackage{hyperref}

\newcommand{\todo}[1]{}

\newcommand{\bfv}[1]{{\bf #1}}

\newcommand{\deriv}[2]{\frac{\text{d}#1}{\text{d}#2}}

\newcommand{\pderivw}[2]{\frac{\partial#1}{\partial#2}}
\newcommand{\expect}[1]{\mathbb{E}\left[#1\right]}
\newcommand{\expectw}[2]{\mathbb{E}_{#1}\left[#2\right]}
\newcommand{\variance}[1]{\mathbb{V}\left[#1\right]}

\newcommand{\estmr}[1]{\hat{#1}}

\usepackage{caption}
\usepackage{subcaption}

\usepackage[accepted]{icml2018}

\icmltitlerunning{Probabilistic Inference for Particle-based Policy Search}

\begin{document}

\twocolumn[ \icmltitle{PIPPS: Flexible Model-Based Policy Search\\
Robust to the Curse of Chaos}

\icmlsetsymbol{equal}{*}

\begin{icmlauthorlist}
\icmlauthor{Paavo Parmas}{oist}
\icmlauthor{Carl Edward Rasmussen}{cam}
\icmlauthor{Jan Peters}{darm,max}
\icmlauthor{Kenji Doya}{oist}
\end{icmlauthorlist}

\icmlaffiliation{oist}{Okinawa Institute of Science and Technology Graduate University, Okinawa, Japan}
\icmlaffiliation{cam}{University of Cambridge, Cambridge, UK}
\icmlaffiliation{darm}{TU Darmstadt, Darmstadt, Germany}
\icmlaffiliation{max}{Max Planck Institute for Intelligent Systems,
 T\"ubingen, Germany}%ü

\icmlcorrespondingauthor{Paavo Parmas}{paavo.parmas@oist.jp}

\icmlkeywords{Machine Learning, Reinforcement Learning, Policy Search, Model-based Reinforcement Learning,
  Gaussian Processes, Stochastic Gradient Estimation, PILCO, Reparameterization,
Likelihood ratio}

\vskip 0.3in
]



\printAffiliationsAndNotice

\begin{abstract}
  Previously, the exploding gradient problem has been explained to be
  central in deep learning and model-based reinforcement learning,
  because it causes numerical issues and instability in
  optimization. Our experiments in model-based reinforcement learning
  imply that the problem is not just a numerical issue, but it may be
  caused by a fundamental chaos-like nature of long chains of
  nonlinear computations. Not only do the magnitudes of the gradients
  become large, the direction of the gradients becomes essentially
  random. We show that reparameterization gradients suffer from the
  problem, while likelihood ratio gradients are robust. Using our
  insights, we develop a model-based policy search framework, {\it
    Probabilistic Inference for Particle-Based Policy Search} (PIPPS),
  which is easily extensible, and allows for almost arbitrary models
  and policies, while simultaneously matching the performance of
  previous data-efficient learning algorithms. Finally, we invent
  the {\it total propagation algorithm}, which efficiently
  computes a union over all pathwise derivative depths during a single
  backwards pass, automatically
  giving greater weight to estimators with lower variance, sometimes
  improving over reparameterization gradients by $10^6$ times.
\end{abstract}

\section{Introduction}
\label{intro}

\begin{figure}[!t]
	\centering
	\begin{subfigure}{.5\columnwidth}
		\includegraphics[width=1\textwidth]{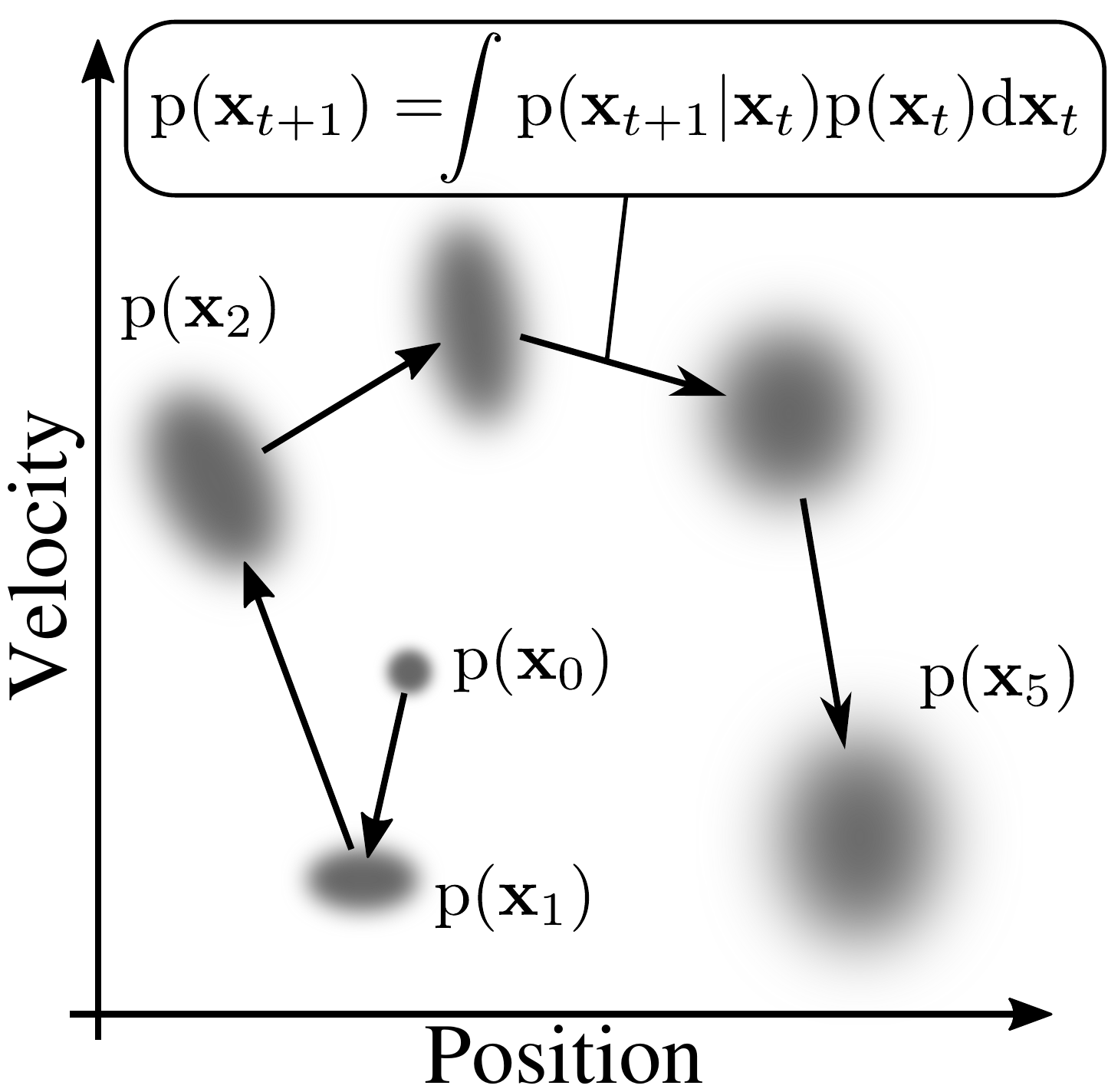}
		\caption{Analytic approximation}
	\end{subfigure}
	\begin{subfigure}{.474\columnwidth}
          \includegraphics[width=1\textwidth]{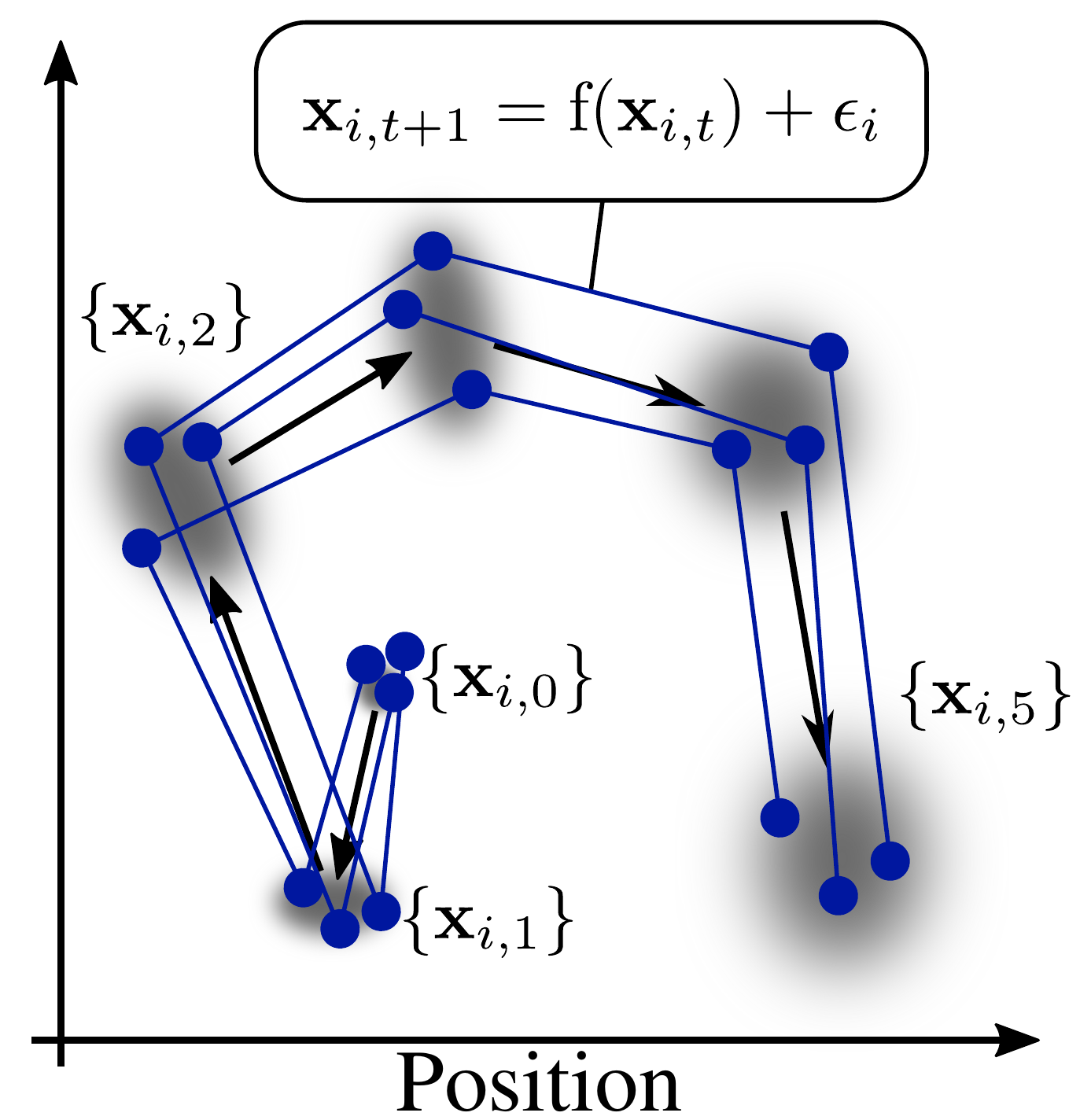}
		\caption{Particle approximation}
        \end{subfigure}
        \caption{One can either perform extensive mathematical
          derivations to analytically predict an approximate
          trajectory distribution, or use a flexible particle approach to predict a stochastic
          approximation to the true trajectory distribution.
          }
        \label{traj}
      \end{figure}

We were motivated by {\it
  Probabilistic Inference for Learning Control} (PILCO) \cite{deisenroth2011pilco}, a model-based reinforcement learning (RL) algorithm which showed impressive results by learning continuous control tasks using several orders of magnitude less data than model-free alternatives. The keys to its success were a
principled approach to model uncertainty, and analytic moment-matching (MM) based
Gaussian approximations of trajectory distributions.

Unfortunately the MM framework is computationally
inflexible. For example, it cannot be used with neural network models. Thus, this work
searches for an alternative flexible method for evaluating
trajectory distributions and gradients.

Particle sampling methods are a general scheme,
which can be applied in practically any setting.  Indeed,
model-free algorithms have previously been successfully used with particle trajectories from the same types of models as used in PILCO
\cite{kupcsik2014gpreps},
\cite{chatzilygeroudis2017blackdrops}. Aiming for better performance,
model-based gradients can be evaluated using the
reparameterization (RP) trick to differentiate through stochasticities.
This approach was previously attempted in the context of PILCO, but surprisingly, it did not
work, with the poor
performance attributed to local minima \cite{andrew}.

Recently, several works have attempted a similar method using Bayesian
neural network dynamics models and the RP trick.
\citet{depeweg2016bbalpha} successfully used this approach to solve
non-standard problems. \citet{gal2016nnpilco}, on the other hand,
found that the direct approach with RP did
not work on the  standard cart-pole swing-up task. 
It is difficult to compare the new approaches to PILCO,
because they simultaneously change multiple aspects of the algorithm:
they switch the model to a more expressive one, then modify the policy
search framework to accommodate this change. In our work, we perform
a shorter step -- we keep everything about PILCO the same, only
changing the framework used for prediction. This approach
allows better explaining how MM helps with learning. We
find that reducing local minima was in fact not the main reason for its
success over particle-based methods. The primary issue was a
hopelessly large gradient variance when using
particles and the RP trick.

We show that the large variance is due to
chaos-like sensitive dependence on initial conditions -- a common
property in calculations involving long chains of nonlinear
mappings. We refer to this
problem as "the curse of chaos". 
Our work suggests that RP gradients and
backpropagation alone are not enough. One either needs methods to
prevent the curse from occurring, or other types of
gradient estimators.

We derive new flexbile gradient estimators, which combine model-based
gradients with the {\it likelihood ratio} (LR) trick
\cite{glynn1990likelihood}, also called REINFORCE
\cite{williams1992reinforce} or the score function
estimator. Our use of LR differs from the typical use in model-free
RL -- instead of sampling with a stochastic policy in the action space,
we use a deterministic policy, but sample with a stochastic model in
the state space. We also develop an importance sampling scheme for use
within a batch of particles. Our estimators obtain accurate gradients,
and allow surpassing the performance of
PILCO.

Our results -- LR gradients perform better than
RP with backpropagation -- are contradictory to recent
work in stochastic variational inference, which suggest that even a
single sample point yields a good gradient estimate using the
RP trick
\cite{kingma2013auto,rezende2014stochasticBP,ruiz2016generalizedRP}. In
our work, not only is a single sample not enough, even millions
of particles would not suffice! In contrast, our new estimators achieve
accurate gradients with a few hundred particles.
As LR gradients are also not perfect, we further invent
the \emph{total propagation algorithm}, which efficiently combines
the best of LR and RP gradients.

\section{Background}
\label{background}
\subsection{Episodic Policy Search}
\label{sec:polsearch}

For a review of policy search methods, see \cite{deisenroth2013policysurvey}.
Consider discrete time systems described by a state vector $\bfv{x}_t$
(the position and velocity of a robot) and the applied action/control
vector $\bfv{u}_t$ (the motor torques). An episode starts by sampling
a state from a fixed initial state distribution
$\bfv{x}_0 \sim p(\bfv{x}_0)$. The policy $\pi_\theta$ determines what
action is applied
$\bfv{u}_t \sim p(\bfv{u}_t) = \pi(\bfv{x}_t;\theta)$. Having applied
an action, the state transitions according to an unknown dynamics
function
$\bfv{x}_{t+1} \sim p(\bfv{x_{t+1}}) = f(\bfv{x}_t, \bfv{u}_t)$.  Both
the policy and the dynamics may be stochastic and nonlinear. Actions
and state transitions are repeated for up to $T$ time-steps, producing
a trajectory $\tau$:
$(\bfv{x}_0,\bfv{u}_0,\bfv{x}_1,\bfv{u}_1,...,\bfv{x}_T)$.  Each
episode is scored according to the return function $G(\tau)$. Often,
the return decomposes into a sum of costs for each time step
$G(\tau) = \sum_{t=0}^Tc(\bfv{x}_t)$, where $c(\bfv{x})$ is the cost
function. The goal is to optimize the policy parameters $\theta$ to
minimize the expected return
$J(\theta) = \expectw{\tau \sim p(\tau;\theta)}{G(\tau)}$. We define
the value $V_h(\bfv{x}) = \expect{\sum_{t=h}^Tc(\bfv{x}_t)}$.

Learning alternates between executing the policy on the system, then
updating $\theta$ to improve the performance on the following
attempts.  Policy gradient methods directly estimate the gradient of
the objective function $\deriv{~}{\theta}J(\theta)$ and use it for
optimization.  Some model-based policy search methods use all of the
data to learn a model of $f$ denoted by $\estmr{f}$, and use it for
"mental rehearsal" between trials to optimize the policy.  Hundreds of
simulated trials can be performed per real trial, greatly increasing
data-efficiency. We utilize the fact that $\estmr{f}$ is
differentiable to obtain better gradient estimators over model-free
algorithms. Importantly, our models are \emph{probabilistic}, and
predict state \emph{distributions}.

\subsection{Stochastic Gradient Estimation}
\label{sec:stocgradest}

Here we explain methods to compute the gradient of the expectation of
an arbitrary function $\phi(x)$ with respect to the parameters of the
sampling distribution
$\deriv{~}{\theta}\expectw{x \sim p(x;\theta)}{\phi(x)}$, e.g.  the
expected return w.r.t. the policy parameters.

{\bf Reparameterization gradient (RP):} Consider sampling from a
univariate Gaussian distribution.  One approach first samples
with zero mean and unit variance
$\epsilon \sim \mathcal{N}(0,1)$, then maps
this point to replicate a sample from the desired distribution
$x = \mu + \sigma\epsilon$.  Now it is straight-forward to
differentiate the output w.r.t. the distribution parameters, namely
$\deriv{x}{\mu} = 1$ and $\deriv{x}{\sigma} = \epsilon$. Averaging
samples of $\deriv{\phi}{x}\deriv{x}{\theta}$ gives an
unbiased estimate of the gradient of the expectation. This is the
RP gradient for a normal distribution. For a
multivariate Gaussian, the Cholesky factor
($L, \text{s.t.} \Sigma = LL^T$) of the covariance matrix can be used
instead of $\sigma$. 
See \cite{rezende2014stochasticBP} for non-Gaussian distributions.

{\bf Likelihood ratio gradient (LR):}
The desired gradient can be written as
$\deriv{~}{\theta}\expectw{x \sim p(x;\theta)}{\phi(x)} = \int
\deriv{p(x;\theta)}{\theta}\phi(x)\text{d}x$. In general, any function
can be integrated by sampling from a distribution $q(x)$ by performing
$\int \phi(x)\text{d}x = \int q(x) \frac{\phi(x)}{q(x)}\text{d}x =
\expectw{x \sim q}{\frac{\phi(x)}{q(x)}}$. The likelihood ratio
gradient picks $q(x) = p(x)$, and directly integrates:
\begin{equation*}
\begin{aligned}
\int \deriv{p(x;\theta)}{\theta}\phi(x)\text{d}x &= \expectw{x \sim
  p}{\frac{\deriv{p(x;\theta)}{\theta}}{p(x)}\phi(x)} \\
%&= \expectw{x \sim p}{\deriv{~}{\theta}\log p(x;\theta)\phi(x)}\\
&= \expectw{x \sim p}{\deriv{\log p(x;\theta)}{\theta}\phi(x)}
\end{aligned}
\end{equation*}
The LR gradient often has a high variance, and has to
be combined with variance reduction techniques known as control
variates \cite{greensmith2004controlvariates}.  A common approach subtracts a constant baseline $b$ from the function values to obtain
the estimator
$\expectw{x \sim p}{\deriv{~}{\theta}\left(\log
    p(x;\theta)\right)(\phi(x)-b)}$.  If $b$ is independent from the
samples, this can greatly reduce the variance without introducing any
bias. In practice, the sample mean
is a good choice $b = \expect{\phi(x)}$. When estimating
the gradient from a batch, one can estimate
leave-one-out baselines for each point to obtain
an unbiased gradient estimator \cite{mnih2016looLR}, i.e.
$b_i = \sum_{j\neq i}^P\phi(x_j)/(P-1)$.

\subsection{Trajectory Gradient Estimation}
\label{sec:trajgrad}
The probability density
$p(\tau) = p(\bfv{x}_0,\bfv{u}_0,\bfv{x}_1,\bfv{u}_1,...,\bfv{x}_T)$
of observing a particular trajectory can be written as
$p(\bfv{x}_0)\pi(\bfv{u}_0|\bfv{x}_0)p(\bfv{x}_1|\bfv{x}_0,
\bfv{u}_0)...p(\bfv{x}_T|\bfv{x}_{T-1},\bfv{u}_{T-1})$.

To use RP gradients, one must know or estimate the dynamics
$p(\bfv{x}_{t+1}|\bfv{x}_t,\bfv{u}_t)$ -- in other words, RP is not
applicable to the model-free case. With a model, a predicted
trajectory can be differentiated by using the chain
rule. 

To use LR gradients, note that $p(\tau)$ is a product, so
$\log p(\tau)$ can be transformed into a sum. Denote
$G_h(\tau) = \sum_{t=h}^Tc(\bfv{x}_t)$. Noting that (1) only the
action distributions depend on the policy parameters, and (2) an
action does not affect costs obtained at previous time steps leads to
the gradient estimator:
$\expect{\sum_{t=0}^{T-1}\left(\deriv{~}{\theta}\log
    \pi(\bfv{u}_t|\bfv{x}_t;\theta)G_{t+1}(\tau)\right)}$

\begin{algorithm}[tb]
  \caption{Analytic moment matching based trajectory prediction and policy evaluation
    (used in PILCO)}
   \label{alg:anal-moment}
\begin{algorithmic}
  \STATE {\bfseries Input:} policy $\pi$ with parameters $\theta$, episode length $T$, initial
  Gaussian state distribution
  $p(\bfv{x}_0)$, cost function $c(\bfv{x})$, learned dynamics model $\estmr{f}$.
  \STATE {\bfseries Requirements:} if the input distribution is Gaussian, can analytically compute
  the expectations and variances of the outputs of $\pi(\bfv{x})$, $\estmr{f}(\bfv{x},\bfv{u})$,
  $c(\bfv{x})$, and differentiate them. 
  \FOR{$t=0$ {\bfseries to} $T-1$}
  \STATE 1. Using $p(\bfv{x}_t)$ and $\pi$ compute a Gaussian approximation to the joint state-action distribution:
  \STATE ~~~~$p(\bfv{\tilde{x}}_t) = \mathcal{N}(\tilde{\mu}_t, \tilde{\Sigma}_t)$, where $\bfv{\tilde{x}}_t=[x^T_t,u^T_t]^T$
  \STATE 2. Using $p(\bfv{\tilde{x}}_t)$ and $\estmr{f}$ compute a Gaussian approximation to the next state
  distribution:
  \STATE ~~~~$p(\bfv{x}_{t+1}) = \mathcal{N}(\mu_{t+1}, \Sigma_{t+1})$
  \STATE 3. Using $p(\bfv{x}_{t+1})$ and $c(\bfv{x})$ compute the expected cost:
  \STATE ~~~~$\expect{c(\bfv{x}_{t+1})}$
  \ENDFOR
  \STATE {\bfseries Gradient computation:}
  $\deriv{~}{\theta}\left(\sum_{t=1}^{T}\expect{c(\bfv{x}_{t})}\right)$ is computed
  analytically during the for-loop by differentiating each computation separately and
  applying the chain rule.
\end{algorithmic}
\end{algorithm}

\subsection{PILCO}

The higher level view of PILCO follows Section~\ref{sec:polsearch} and
the policy gradient evaluation is detailed in
Algorithm~\ref{alg:anal-moment}.

\subsubsection{Probabilistic Dynamics Model}
We
follow the original PILCO, which uses Gaussian process \cite{gpbook}
dynamics models to predict the change in state from one time step to
the next, i.e.
$p(\Delta x_{t+1}^a) = \mathcal{GP}(\bfv{x}_t, \bfv{u}_t)$, where
$\bfv{x}\in\mathbb{R}^D$, $\bfv{u}\in\mathbb{R}^F$ and
$\Delta x_{t+1}^a = x_{t+1}^a-x_{t}^a$.  A separate Gaussian process
is learned for each dimension $a$.  We use a squared exponential
covariance function
$k_a(\bfv{\tilde{x}},\bfv{\tilde{x}'}) = s_a^2\exp(-(\bfv{\tilde{x}} -
\bfv{\tilde{x}'})^T\Lambda_a^{-1}(\bfv{\tilde{x}} -
\bfv{\tilde{x}'}))$, where $s_a$ and
$\Lambda = \text{diag}([l_{a1}, l_{a2}, ..., l_{a{D+F}}])$ are the
function variance and length scale hyperparameters respectively. We
use a Gaussian likelihood function with a noise hyperparameter
$\sigma_n$.  The hyperparameters are trained to maximize the marginal
likelihood.  When sampling from these models, the prediction has the
form $y = \estmr{f}(\bfv{x}) + \epsilon$, where
$\epsilon \sim \mathcal{N}\left(0,\sigma_f^2(\bfv{x}) +
  \sigma_n^2\right)$. Here $\sigma_f^2$ represents the model
uncertainty, and is caused by a lack of data in a region, while
$\sigma_n^2$ is the learned inherent model noise. The learned
model noise is not necessarily the same as the real observation
noise $\sigma_o^2$ in the system. In particular, the latent state
is not modeled and the system is approximated by predicting
the next observation given the current
observation. Moreover, there is an additional source of the variance in the
trajectory -- with different start locations, the trajectory will differ.

\subsubsection{Moment matching Prediction}
In general, when a Gaussian distribution is mapped through a nonlinear
function, the output is intractable and non-Gaussian; however, in some
cases one can analytically evaluate the moments of the output
distribution. Moment-matching (MM) approximates the output
distribution as Gaussian by matching the mean and variance with the
true moments. Note that even though the state-dimensions are modelled
with separate functions $\estmr{f}_a$, MM is performed jointly, and
the state distributions can include covariances.

\section{Particle Model-Based Policy Search}
\label{parmops}

\subsection{Particle Prediction}

In general, particle trajectory predictions are simple --
predict at all particle locations, sample from the output distributions, repeat.
However, we also compare to a scheme based on Gaussian resampling (GR), used by \citet{gal2016nnpilco} to apply PILCO with neural network dynamics models.

{\bf Gaussian resampling (GR):} MM can be stochastically
replicated. At each time step, the mean
$\estmr{\mu} = \sum_{i=1}^P\bfv{x}_i/P$ and covariance
$\estmr{\Sigma} =
\sum_{i=1}^P(\bfv{x}_i-\estmr{\mu})(\bfv{x}_i-\estmr{\mu})^T/(P-1)$ of
the particles are estimated. Then the particles are resampled from the
fitted distribution
$x'_i \sim \estmr{\mu} + L\bfv{z}_i ~|~ \bfv{z}_i \sim
\mathcal{N}(0,I)$, where $L$ is the Cholesky factor of
$\estmr{\Sigma}$. One can differentiate this
resampling operation by using RP. Obtaining the gradient
$\text{d}{L}/\text{d}{\estmr{\Sigma}}$ is non-trivial, but \cite{murray2016choldiv} presents an overview. We use the provided
symbolic expression.

\subsection{Hybrid Gradient Estimation Techniques}
  
In our case, RP gradients can be used. However, surprisingly
they were hopelessly inaccurate (see Figure~\ref{fig:gradvar}). To
solve the problem, we derived new gradient estimators which combine model derivatives with
LR gradients. In particular, our approach allowed for
within batch importance sampling to increase sample efficiency.

{\bf Model-based LR:} As in Section~\ref{sec:trajgrad}, one can
write the distribution over predicted trajectories as
$p(\tau) =
p(\bfv{x}_0)\pi(\bfv{u}_0|\bfv{x}_0)\estmr{f}(\bfv{x}_1|\bfv{x}_0,
\bfv{u}_0)...\estmr{f}(\bfv{x}_T|\bfv{x}_{T-1},\bfv{u}_{T-1})$.  With
deterministic policies, the model and policy can be combined:
$p(\bfv{x}_{t+1}|\bfv{x}_t) = \estmr{f}(\bfv{x}_{t+1}|\bfv{x}_t,
\pi(\bfv{x}_t;\theta))$, which is differentiable
$\deriv{p_{t+1}}{\theta} =
\deriv{p_{t+1}}{\bfv{u}_t}\deriv{\bfv{u}_t}{\theta}$. A model-based
gradient derives:
$\expect{\sum_{t=0}^{T-1}\left(\deriv{~}{\theta}\log
    p(\bfv{x}_{t+1}|\bfv{x}_t;\theta)(G_{t+1}(\bfv{x}_{t+1}) - b_{t+1})\right)}$

{\bf Batch Importance Weighted LR (BIW-LR):} We use parallel computation,
and sample multiple particles simultaneously. The
state distribution is represented as a mixture distribution
$q(\bfv{x}_{t+1}) = \sum_{i=1}^Pp(\bfv{x}_{t+1}|\bfv{x}_{i,t};\theta)/P$.
Analogously to the derivation of LR in Section~\ref{sec:stocgradest},
one can derive a lower variance estimator with importance sampling within the batch
 for each time step:
$
\sum_{i=1}^P\sum_{j=1}^P
\left(\frac{\text{d}{p(\bfv{x}_{j,t+1}|\bfv{x}_{i,t};\theta)}/\text{d}{\theta}}
  {\sum_{k=1}^Pp(\bfv{x}_{j,t+1}|\bfv{x}_{k,t})}(G_{t+1}(\bfv{x}_{j,t+1}) - b_{i,t+1})\right)/P
$

We choose to estimate a leave-one-out mean of the returns by
normalized importance sampling with the equation:
$b_{i,t+1} = \left(\sum_{j\neq i}^P
  c_{j,t+1}G_{t+1}(\bfv{x}_{j,t+1})\right)/\sum_{j\neq i}^P
c_{j,t+1}$, where
$c_{j,t+1} = p(\bfv{x}_{j,t+1}|\bfv{x}_{i,t})/
\sum_{k=1}^Pp(\bfv{x}_{j,t+1}|\bfv{x}_{k,t})$. Without
normalizing, a large variance of the baseline
estimation leads to poor LR gradients.  Note that we
compute $P$ baselines for each time-step, whereas there are $P^2$
components in the gradient estimator. To obtain a true unbiased
gradient, one should compute $P^2$ leave-one-out baselines -- one for
each particle for each mixture component of the distribution. The
paper contains evaluations only with the baseline presented here -- we
found that it already removes most of the bias.

\begin{algorithm}[tb]
  \caption{Total Propagation Algorithm\\
    (used in PIPPS for evaluating the gradient)}
   \label{alg:total-prop}
\begin{algorithmic}
  \STATE This algorithm provides an efficient method to fuse LR and RP
  gradients by combining ideas from filtering and back-propagation.
  The algorithm is explained here with reference to our policy search
  framework.  \STATE {\bfseries Forward pass:} Compute a set of
  particle trajectories.
  \STATE {\bfseries Backward pass:}\\
  \STATE {\bfseries Initialise:}
  $\deriv{G_{T+1}}{\zeta_{T+1}} = \bfv{0}$,
  $\deriv{J}{\theta} = \bfv{0}$, $G_{T+1} = 0$ where $\zeta$ are the
  distribution parameters, e.g. $\mu$ and $\sigma$.
  \FOR{$t=T$ {\bfseries to} $1$}
  \FOR{{\bfseries each} particle $i$}
  \STATE $G_{i,t} = G_{i,t+1} + c_{i,t}$, where $c_t$ is the cost at
  time t.  \STATE
  $\deriv{\bfv{\zeta}_{i,t+1}}{\bfv{x}_{i,t}} =
  \pderivw{\bfv{\zeta}_{i,t+1}}{\bfv{x}_{i,t}} +
  \deriv{\bfv{\zeta}_{i,t+1}}{\bfv{u}_{i,t}}\deriv{\bfv{u}_{i,t}}{\bfv{x}_{i,t}}$
  \STATE
  $\deriv{G_{i,t}^{RP}}{\zeta_{i,t}} =
  (\deriv{G_{i,t+1}}{\zeta_{i,t+1}}
  \deriv{\zeta_{i,t+1}}{\bfv{x}_{i,t}} +
  \deriv{c_{i,t}}{\bfv{x}_{i,t}})\deriv{\bfv{x}_{i,t}}{\zeta_{i,t}}$
  \STATE
  $\deriv{G_{i,t}^{LR}}{\zeta_{i,t}} =
  (G_{i,t}-b_{i,t})\deriv{\log p(\bfv{x}_{i,t})}{\zeta_{i,t}}$
  \STATE
  $\deriv{G_{i,t}^{RP}}{\theta} = \deriv{G_{i,t}^{RP}}
  {\zeta_{i,t}}\deriv{\zeta_{i,t}}{\bfv{u}_{i,t-1}}\deriv{\bfv{u}_{i,t-1}}{\theta}$
  \STATE
  $\deriv{G_{i,t}^{LR}}{\theta} = \deriv{G_{i,t}^{LR}}
  {\zeta_{i,t}}\deriv{\zeta_{i,t}}{\bfv{u}_{i,t-1}}\deriv{\bfv{u}_{i,t-1}}{\theta}$
  \ENDFOR
  \STATE
  $\sigma^2_{RP} = \text{trace}(\variance{\deriv{G_{i,t}^{RP}}{\theta}})$,
  $\sigma^2_{LR} = \text{trace}(\variance{\deriv{G_{i,t}^{LR}}{\theta}})$
  \STATE
  $k_{LR} = 1/\left(1+\frac{\sigma^2_{LR}}{\sigma^2_{RP}}\right)$
  \STATE
  $\deriv{J}{\theta} = \deriv{J}{\theta} + k_{LR}\frac{1}{P}\sum_i^P \deriv{G_{i,t}^{LR}}{\theta} +
  (1-k_{LR})\frac{1}{P}\sum_i^{P} \deriv{G_{i,t}^{RP}}{\theta}$
  \FOR{{\bfseries each} particle $i$}
  \STATE
  $\deriv{G_{i,t}}{\zeta_{i,t}} = k_{LR}\deriv{G_{i,t}^{LR}}{\zeta_{i,t}} +
  (1-k_{LR})\deriv{G_{i,t}^{RP}}{\zeta_{i,t}}$
  \ENDFOR  
  \ENDFOR
\end{algorithmic}
\end{algorithm}

{\bf RP/LR weighted average:} The bulk of the computation is spent on
the $\text{d}p(\bfv{x}_{t+1}|\bfv{x_t};\theta)/\text{d}\theta$
terms. These terms are needed for both LR and RP gradients, so there
is no penalty to combining both estimators. A well known statistics
result states that for independent estimators, an optimal weighted
average estimate is achieved if the weights are proportional to the inverse
variance,
 i.e.
$\mu = \mu_{LR}k_{LR} + \mu_{RP}k_{RP}$, where
$k_{LR} = \estmr{\sigma}_{LR}^{-2}/(\estmr{\sigma}_{LR}^{-2} +
\estmr{\sigma}_{RP}^{-2})$ and $k_{RP} = 1-k_{LR}$.

A naive combination scheme would compute the gradient separately for
the whole trajectory for both estimators, then combine them; however,
this approach neglects the opportunity to use reparameterization
gradients through shorter sections of the trajectory to obtain better
gradient estimates. Our new {\it total propagation algorithm} (TP)
goes beyond the naive method. TP uses a single backward pass to
compute a union over all possible RP depths, automatically giving
greater weight to estimators with lower variance.

A description is provided in Algorithm~\ref{alg:total-prop}. At each
backward step, it evaluates the gradient w.r.t. the policy parameters
using both the LR and RP methods.  It evaluates a ratio based on the
variances in \emph{policy parameter space} -- this variance is
proportional to the variance of the policy gradient estimator.
The gradients are combined, and a "best"
estimate in \emph{distribution parameter space} is passed to the
previous time step.  In the
algorithm, the $\mathbb{V}$ operator takes the sample variance of
gradient estimates from different particles; however, other variance
estimation schemes could also be considered: one could estimate
variances from moving averages of the magnitude of the gradient, use a
different statistical estimator for the variance, use only a subset of
policy parameters, etc.
The algorithm is not limited to RL problems, but
is applicable to general stochastic computation graphs
\cite{schulman2015stocgraph}, and could be used for
training probabilistic models, stochastic neural networks, etc.

\subsection{Policy Optimization}

{\bf RMSprop-like stochastic gradient descent:} We use an algorithm
motivated by RMSprop \cite{tieleman2012rmsprop}.
RMSprop normalizes its SGD steps by utilizing a running
average of the square of the gradients.
In our case, since our batch sizes were large,
we directly estimate the expectation of the square from the batch by
$z =\expect{g^2} = \expect{g}^2 + \variance{g}$, where $g$ is the
gradient. We use the variance of the mean, i.e. $\variance{g}$ is the
variance divided by the number of particles $P$. The gradient step
becomes $g/\sqrt{z}$. We use momentum with the parameter
$\gamma$. The full update equations become:
\begin{equation*}
  \begin{aligned}
    m &\leftarrow \gamma m +  g/\sqrt{\expect{g}^2 + \variance{g}}\\
    \theta &\leftarrow \theta - \alpha m
  \end{aligned}
\end{equation*}

{\bf Deterministic optimizer:} The random number seed can be fixed to
turn a stochastic problem deterministic, also known as the PEGASUS
trick \cite{ng2000pegasus}. With a fixed seed, the RP gradient is an
exact gradient of the objective, and quasi-Newton
optimizers, such as BFGS can be used.

\section{Experiments}
\label{exps}

We performed experiments with two purposes: 1. To
explain why RP gradients
are not sufficient (Section~\ref{valueland}). 2. To show that our
newly developed methods can match up to PILCO in terms of learning
efficiency (Section~\ref{learnexp}).

\subsection{Plotting the Value Landscape}
\label{valueland}

We perturb the policy parameters $\theta$ in a randomly chosen fixed direction, and plot both the objective function,
and the magnitude of the projected gradient as a function of $\Delta \theta$. The results of this experiment are
perhaps the most striking component of our paper, and motivated the term "the curse of chaos". Refer to Figure~\ref{curseofchaos}
for the results, and to Section~\ref{sec:curse} for
a full explanation.

The plots were generated in the nonlinear cart-pole task, using similar settings as explained in
Section~\ref{learnexp}.  We used 1000 particles, and while keeping the
random number seed fixed to demonstrate that the high variance in
\ref{fig:gradvar} is not caused by randomness, but by a chaos-like
property of the system. The confidence intervals were estimated by
$Var/P$ where $Var$ is the sample variance, and $P$ is the number of particles. In
Section~\ref{sec:gradvareval} we plot the dependence of the variance
on $P$ using a more principled approach.

\subsubsection{Explaining the Curse of Chaos}
\label{sec:curse}

\begin{figure*}[!t]
        \centering
	\begin{subfigure}{.3\textwidth}
		\includegraphics[width=\textwidth]{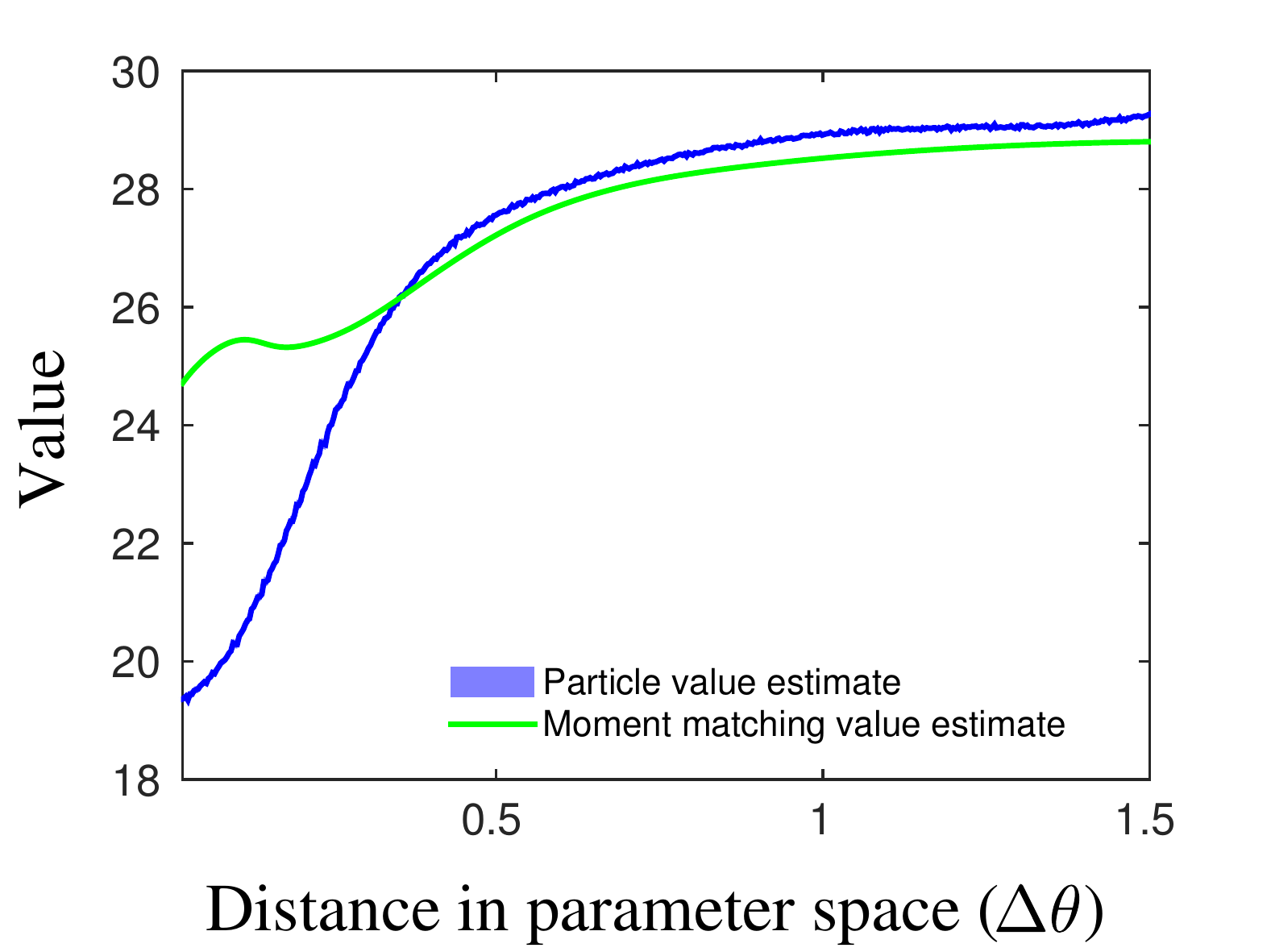}
		\caption{}
          \label{obj}
        \end{subfigure}
%%%%%%%%%%%%%%
	\begin{subfigure}{.3\textwidth}
		\includegraphics[width=\textwidth]{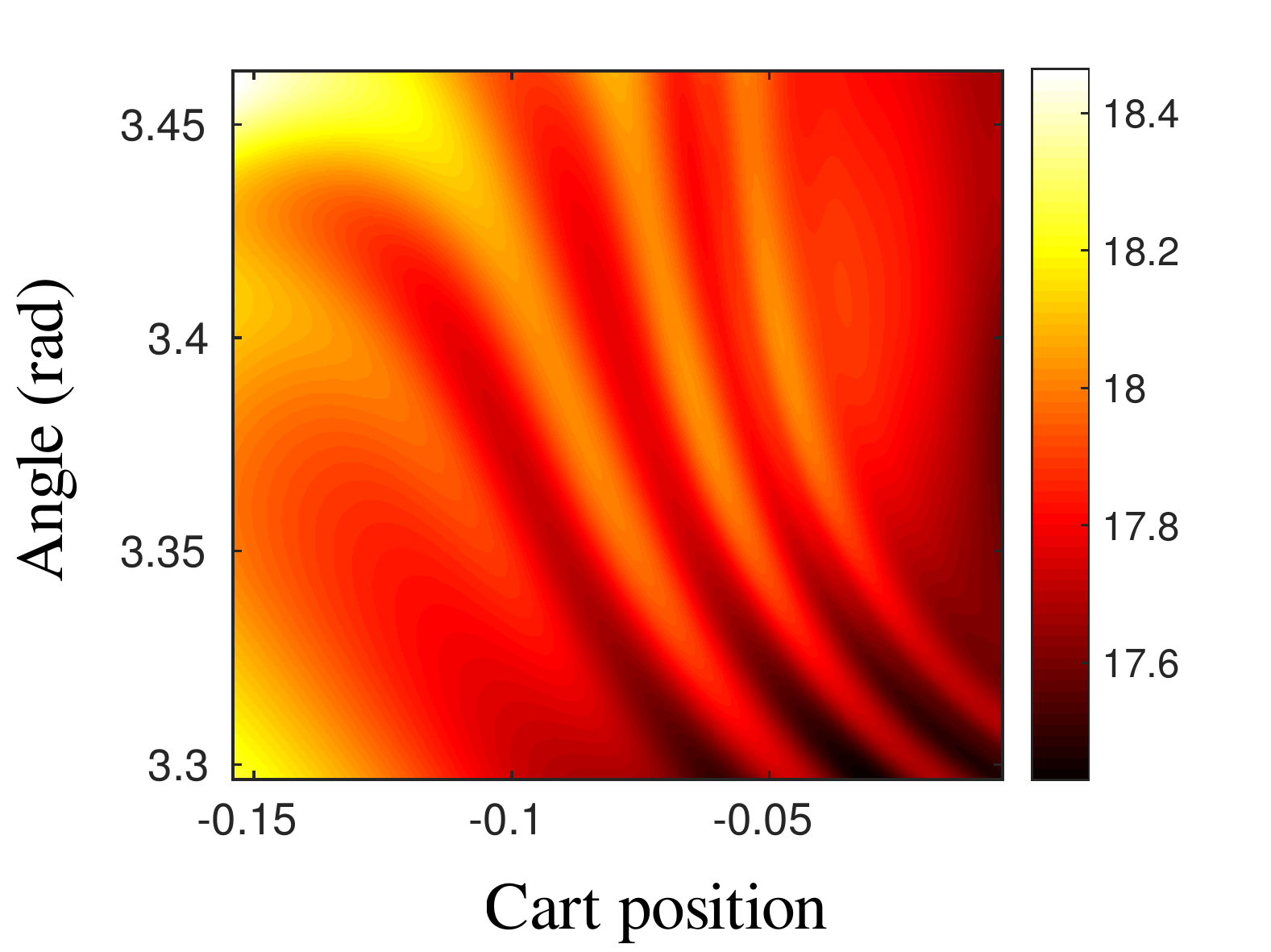}
		\caption{Value at $\text{p}(\bfv{x}_1)$ and
                  $\Delta\theta = 0$}
          \label{lhs}
	\end{subfigure}
%%%%%%%%%%%%%%
	\begin{subfigure}{.3\textwidth}
		\includegraphics[width=\textwidth]{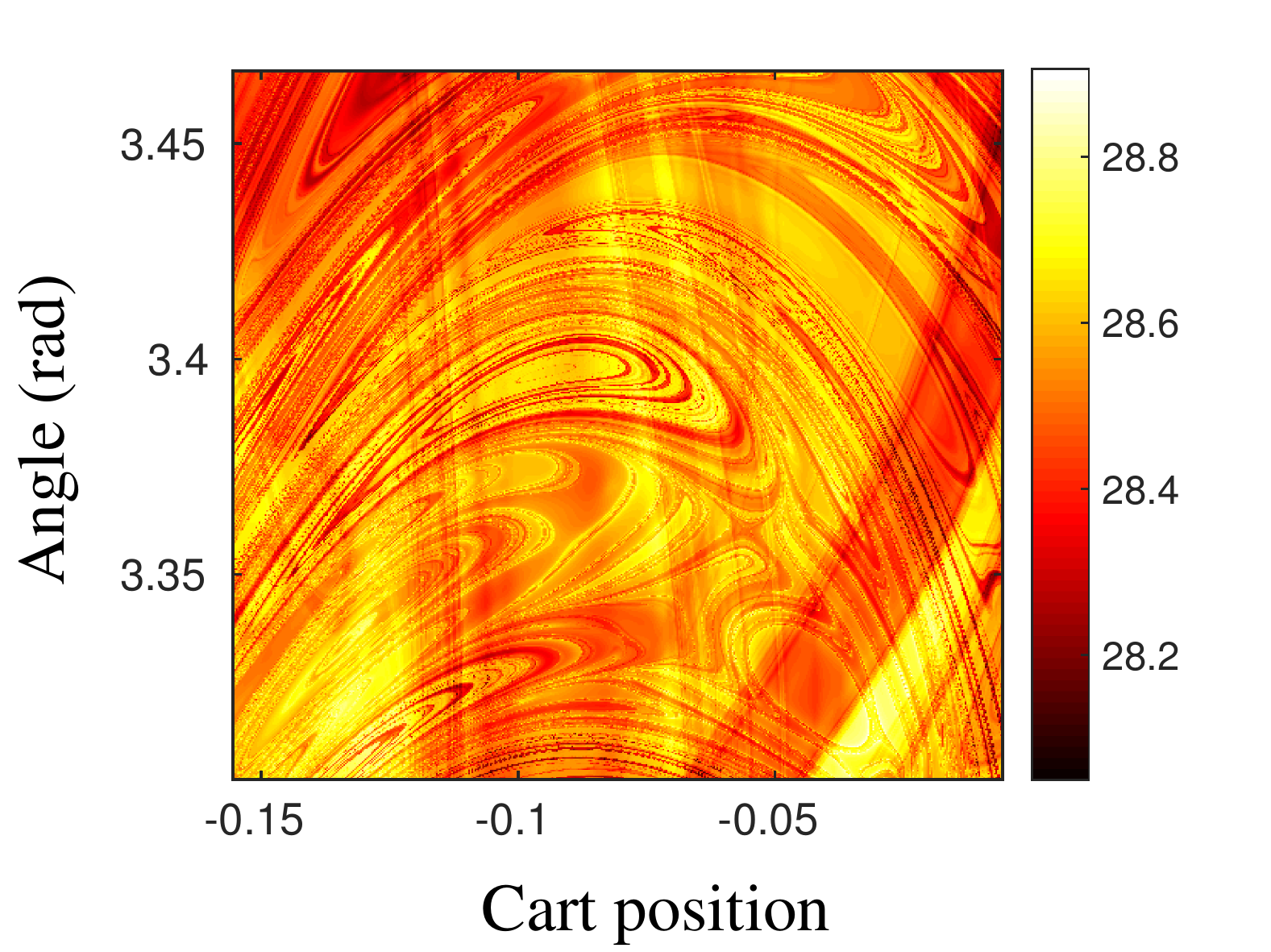}
		\caption{Value at $\text{p}(\bfv{x}_1)$ and
                  $\Delta\theta = 1.5$}
          \label{rhs}
        \end{subfigure}\\
        \begin{subfigure}{.3\textwidth}
		\includegraphics[width=\textwidth]{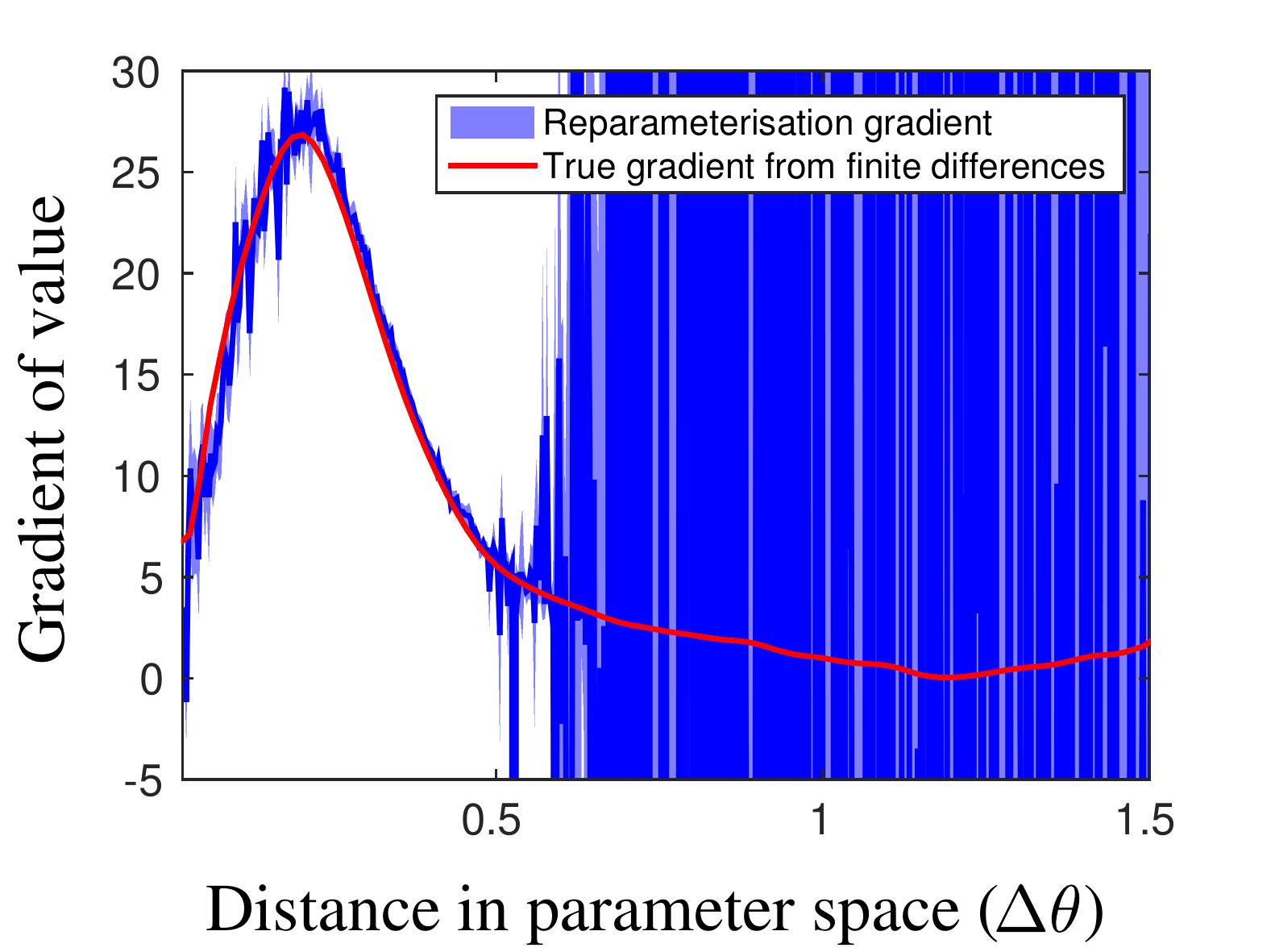}
		\caption{$\text{RP}_{\text{FS}}$ (Exact gradient of value in \ref{obj})}
                \label{fig:gradvar}
	\end{subfigure}
%%%%%%%%%%%%%%
	\begin{subfigure}{.3\textwidth}
		\includegraphics[width=\textwidth]{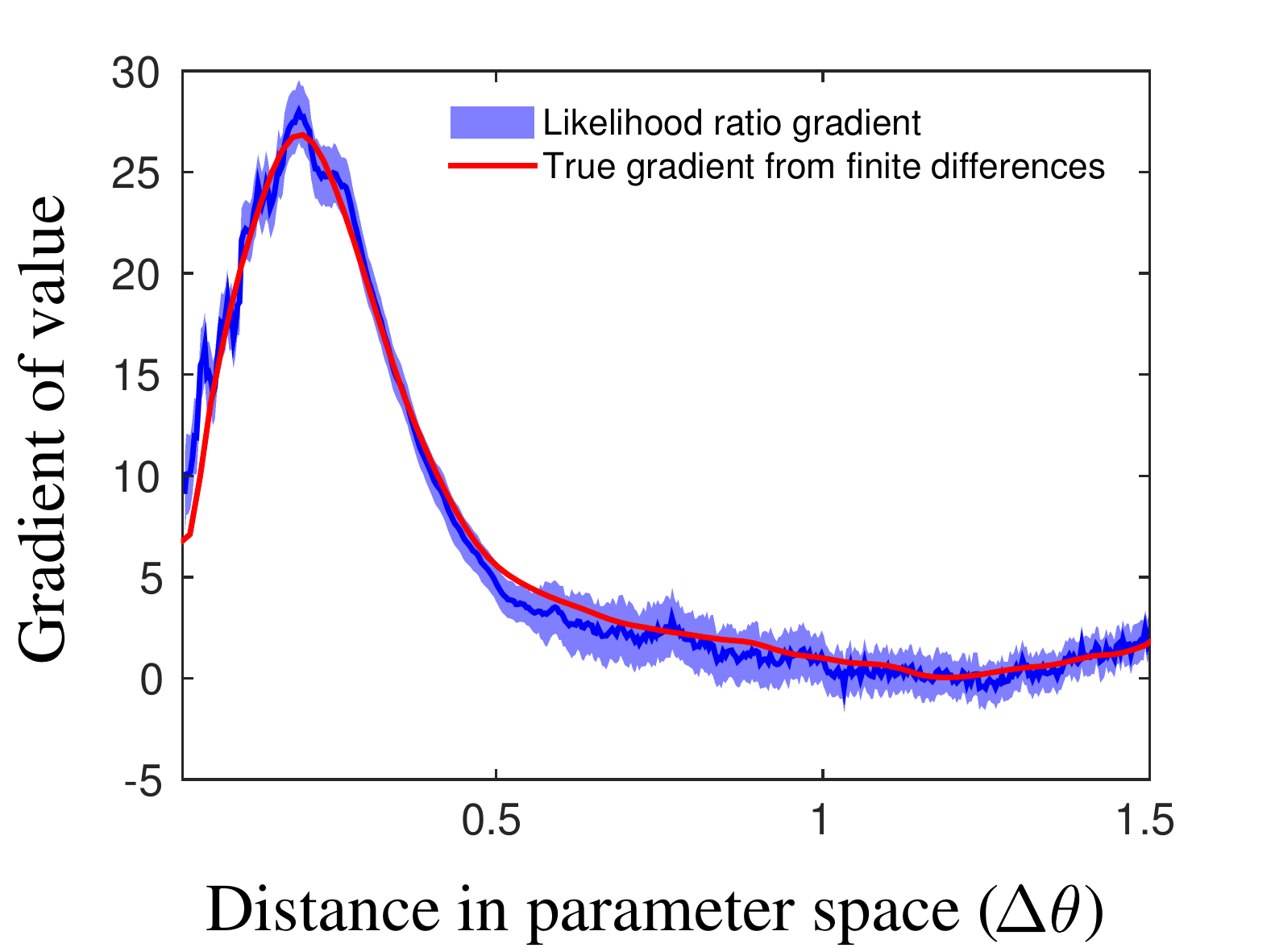}
		\caption{LR}
                \label{lrvar}
	\end{subfigure}
%%%%%%%%%%%%%%
	\begin{subfigure}{.3\textwidth}
		\includegraphics[width=\textwidth]{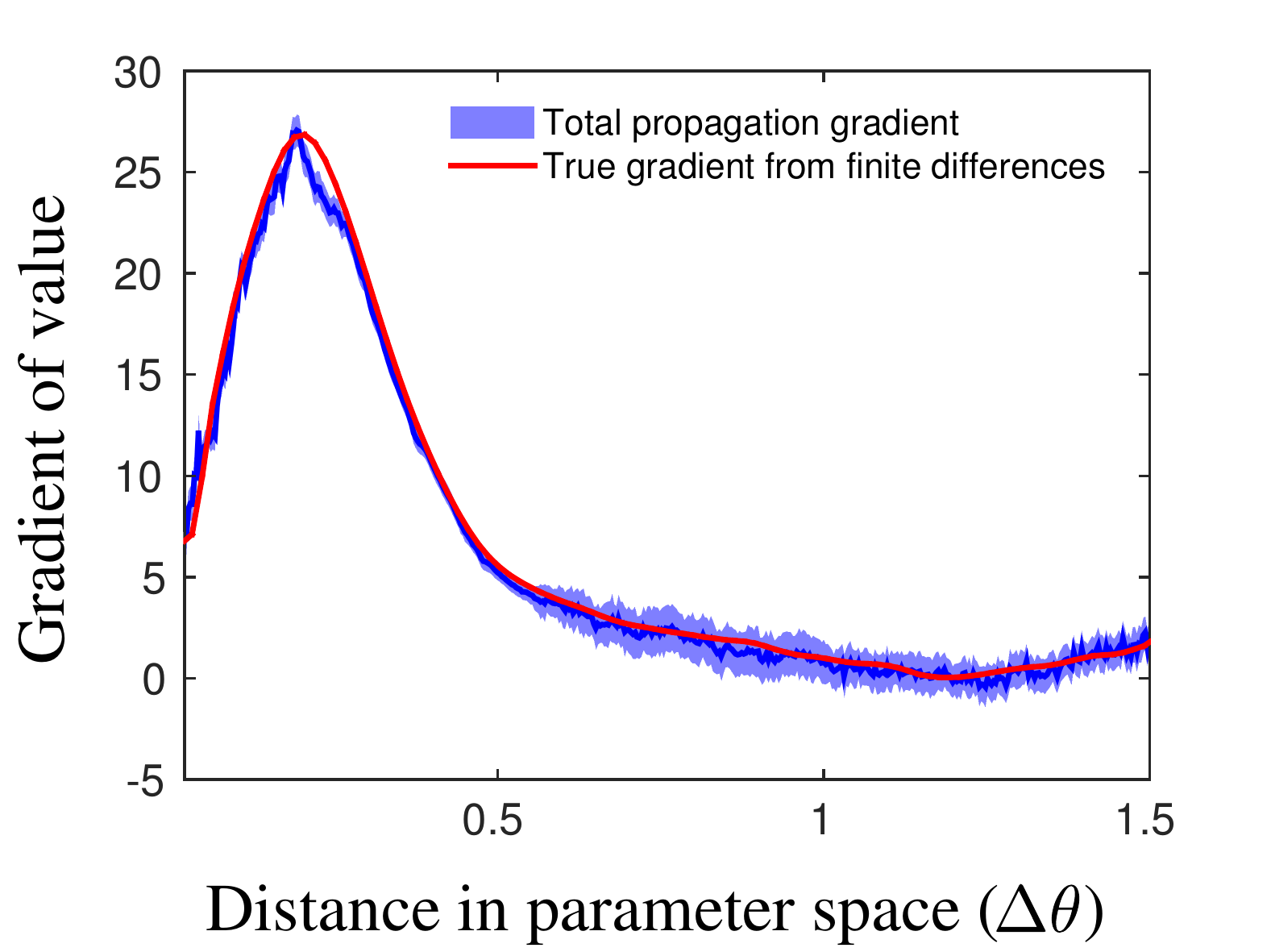}
		\caption{TP}
                \label{TPvar}
        \end{subfigure}
	\caption{To illustrate the behavior of the gradient estimators
          in the cart-pole task, we fix the random number seed, vary
          the policy parameters $\theta$ in a fixed direction, and
          plot a 95\% confidence interval for both the objective
          (\ref{obj}) and the magnitude of the gradient estimators in
          this direction (\ref{fig:gradvar},\ref{lrvar},\ref{TPvar}).
          Figure~\ref{fig:gradvar} shows that reparameterization
          gradients suffer from the curse of chaos, which can cause
          the gradient variance to explode. Our new estimators in
          Figures~\ref{lrvar} and \ref{TPvar} are robust to this
          issue. Figures~\ref{lhs} and \ref{rhs} correspond to the
          parameter settings at the left and right side of the other
          plots, and explain the reason for the exploding gradient
          variance. Refer to Section~\ref{sec:curse} for a full
          explanation. The confidence interval of the particle value
          estimate in (\ref{obj}) is so tight, that the line is
          thicker, but note that a smooth line would fit within the
          error bounds. Moreover, note that having fixed the random
          number seed causes a bias.  }
          \label{curseofchaos}
\end{figure*}

Figure~\ref{fig:gradvar} contains a peculiar result where the RP
gradient behaves well for some regions of the parameters $\theta$,
but when $\theta$
is perturbed, a phase-transition-like change causes the
variance to explode. The variance at $\Delta\theta = 1.5$ is
$\sim$$4\times10^5$ times larger than at $\Delta\theta =
0$, meaning that $\sim$$4\times10^8$ particles would be
necessary for the RP gradient to become accurate in that
region. For practical purposes, optimizing with
the RP gradient would lead to a simple random walk.

Since the seed was fixed, the RP gradient in
\ref{fig:gradvar} is an \emph{exact} gradient of the value in
\ref{obj}. Therefore, there is an infinitesimal
deterministic "noise" at the right of
\ref{obj}. The value averaged across 1000 particles is though not the
true objective -- that would require averaging an
infinite number of particles. When averaging an infinite amount of
particles, is there still a "noise", or does the
function become smooth?

Our new gradient estimators in Figures~\ref{lrvar} and \ref{TPvar}
suggest that the true objective is indeed smooth. To provide more
evidence, we estimated the magnitude of the gradient from finite
differences of the value in \ref{obj} using a sufficiently large
perturbation in $\theta$, such that the "noise" is ignored.  The fact
that two separate approaches agree -- one which varies
the policy parameters $\theta$, and another which keeps $\theta$
fixed, but estimates the gradient from the trajectories -- 
provides convincing evidence that the true objective is smooth.

Figures~\ref{lhs} and \ref{rhs} explain the reason for the explosion
of the variance when using RP gradients
(\ref{fig:gradvar}). Figure~\ref{lhs} corresponds to the leftmost
parameter setting ($\Delta\theta=0$); Figure~\ref{rhs} corresponds to
the rightmost parameter setting ($\Delta\theta=1.5$). The plots show
how the value $V(\bfv{x};\theta)$ (the remaining cumulative cost)
varies as a function of the state position $\bfv{x}$. Note that
because the random number seed is fixed, the value $V$ is the same as
the remaining return $G$. This definition differs from the typical
value function, which averages the return over an infinite amount of
particles. The figures were created by predicting the trajectory at
each point for 4 particles with different fixed seeds, then averaging
the costs of the trajectories. We chose to predict 4 particles after
trying 1 particle, for which the value appeared to include a step-like
part, but was otherwise less interesting than the current figure. As
the average value of the 4 particles is erratic, at least one of the 4
particles must have a highly erratic value in the shown region.

The boxes (\ref{lhs}, \ref{rhs}) are centered at the mean
 prediction from the center of the initial state
distribution (if unclear, consider Figure~\ref{traj} with
$p(\bfv{x}_0)$ as a point mass, then $p(\bfv{x}_1)$ depicts the
location of the box).  The
axes on the boxes are slightly different, because when $\theta$ is changed, the predicted location
$p(\bfv{x}_1;\theta)$ changes. The side lengths correspond to 4
standard deviations of the Gaussian distributions
$p(\bfv{x}_1;\theta)$. The velocities were kept fixed at the mean
values.

RP estimates
$\deriv{}{\theta}\int p(\bfv{x}_1;\theta)V(\bfv{x}_1)\text{d}\bfv{x}$. It samples points inside the box, computes the gradient
$\pderivw{V}{\theta} = \pderivw{V}{x}\deriv{x}{\theta}$ and
averages the samples together\footnote{Note that the same evaluation
  of the value gradient has to be performed at subsequent time-steps,
  and in practice the sum is evaluated simultaneously using
  backpropagation, but we ignore this for the purpose of the
  explanation.}. In Figure~\ref{rhs} finding the gradient
of the expectation by differentiating $V$ is completely hopeless.
In contrast, the LR gradient
(\ref{lrvar}) only uses the value $V$, not its derivative, and does
not suffer from this problem. 

Finally, even though we do not show the plotted value and
 gradient for the Gaussian resampling case,
both of these were smooth functions for a fixed random seed. Thus,
resampling also beats the curse of chaos.

\subsubsection{Policy Gradient Variance Evaluation}
\label{sec:gradvareval}
\begin{figure}
	\centering
	\begin{subfigure}{.49\columnwidth}
		\includegraphics[width=1\textwidth]{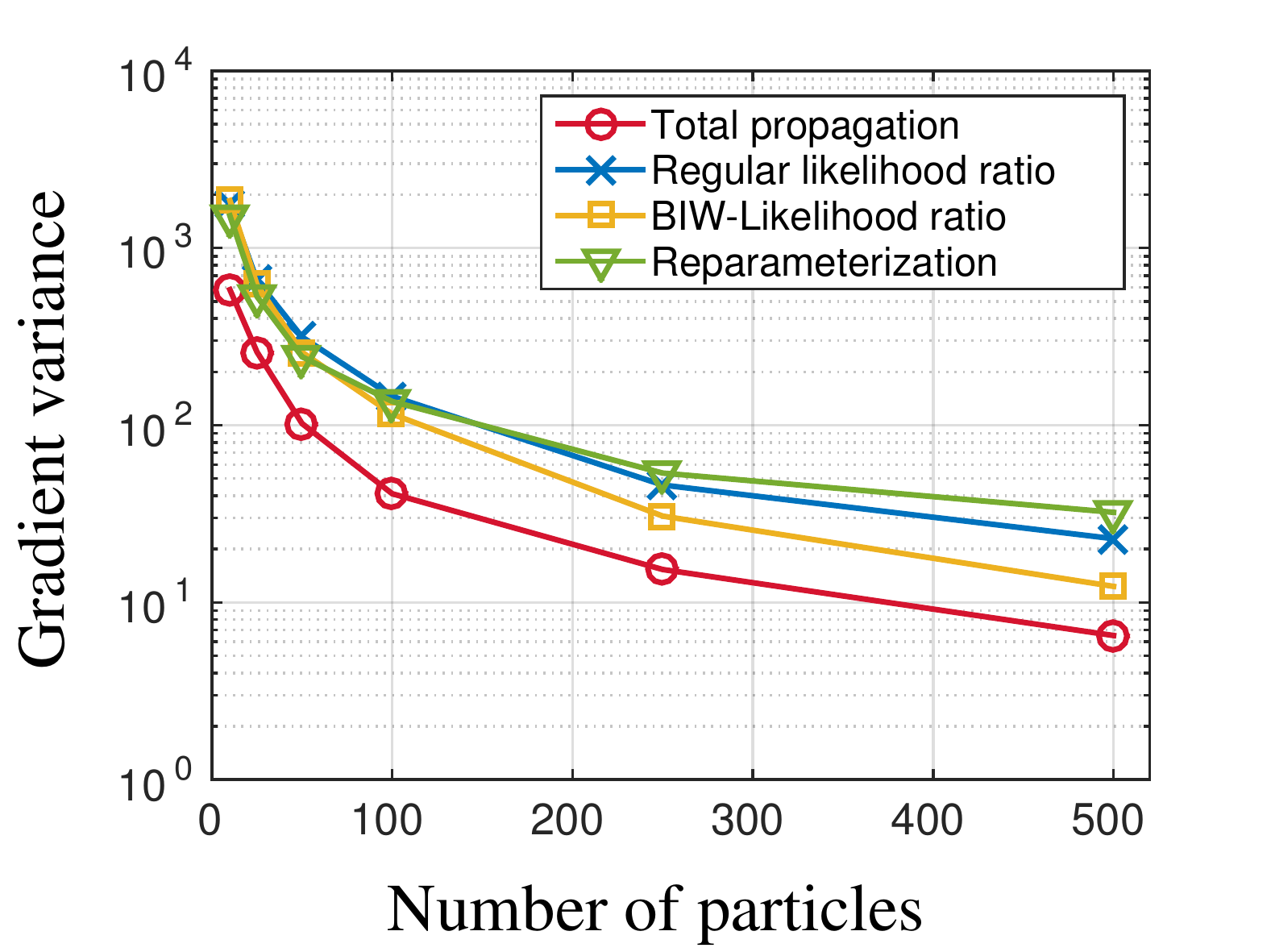}
		\caption{$\Delta\theta = 0$}
	\end{subfigure}
	\begin{subfigure}{.49\columnwidth}
          \includegraphics[width=1\textwidth]{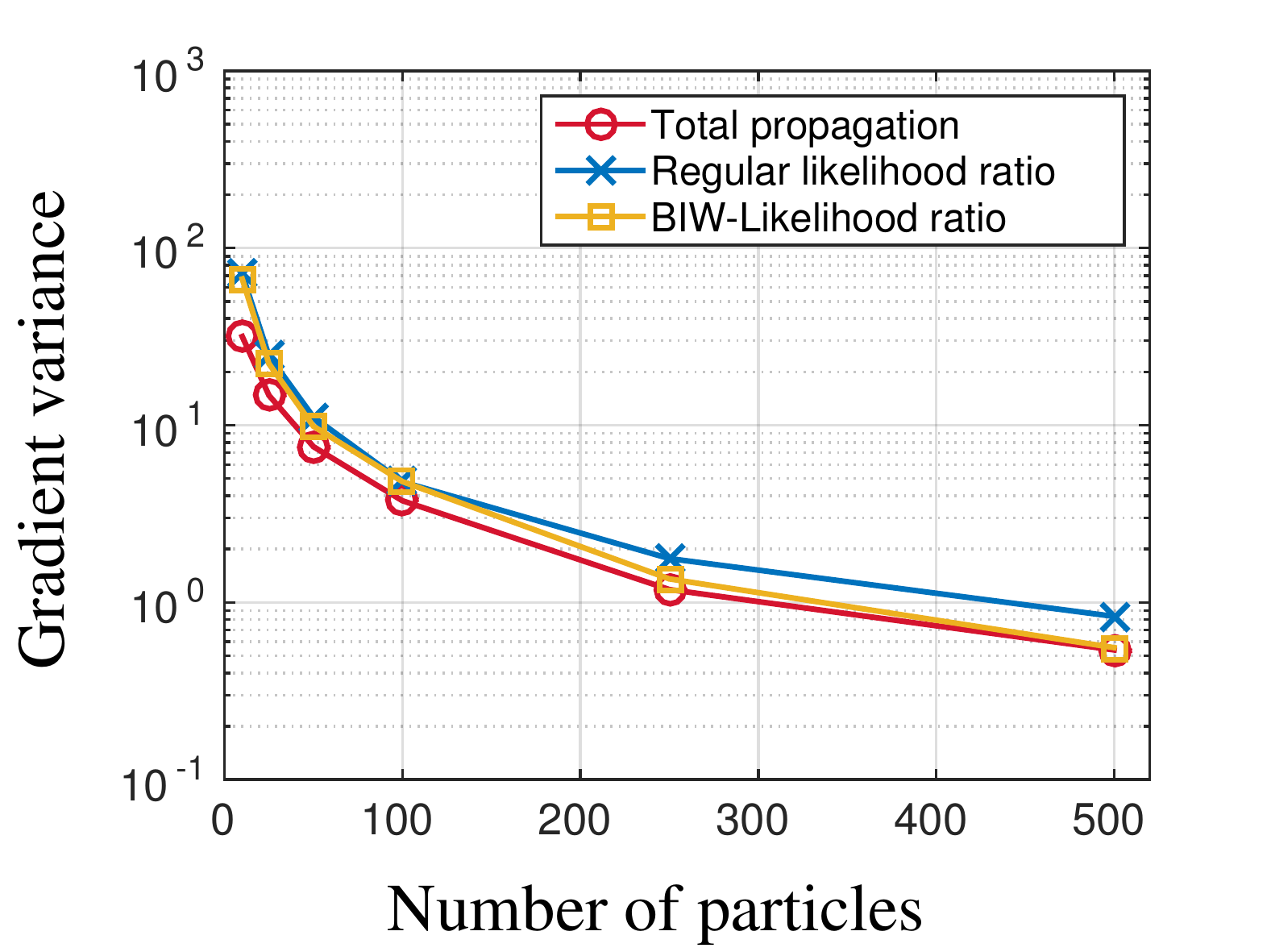}
          \caption{$\Delta\theta = 1.5$}
          \label{rhsvar}
        \end{subfigure}
        \caption{Computed variances corresponding to plots in
          Figure~\ref{curseofchaos}.}
        \label{variancecomp}
\end{figure}

In Figure~\ref{variancecomp}, we plot how the variance of the gradient
estimators at $\Delta\theta = 0$ and $\Delta\theta = 1.5$ depends on
the number of particles $P$. The variance was computed by repeatedly sampling
the estimator for a large number of times and calculating the variance from the set of evaluations.
We compare RP, TP as well as LR gradients both with
and without batch importance weighting (BIW) to show that our importance
sampling scheme reduces the variance. We used
the importance sampled baseline; in practice the regular LR gradient
would use a simpler baseline, and have even higher
variance. The RP gradient is omitted from
\ref{rhsvar}, because the variance was between $10^8$-$10^{15}$. The
TP gradient combined the BIW-LR and RP gradients.

The results confirm that BIW
significantly reduces the variance. Moreover, our TP
algorithm was the best. Importantly, in \ref{rhsvar},
even though the variance of the RP gradient for the full trajectory is
over $10^6$ larger than the other estimators, TP utilizes shorter path-length RP gradients to
obtain 10-50\% reduction in variance for 250 particles and fewer.

\subsection{Learning Experiments}
\label{learnexp}

We compare PILCO in episodic learning tasks to the following particle-based methods:
RP, RP with a fixed seed
(RP$_{\text{FS}}$), Gaussian resampling (GR), GR with
a fixed seed (GR$_{\text{FS}}$), model-based batch importance weighted
likelihood ratio (LR) and total propagation (TP). Moreover, we
evaluate two variations of the particle predictions:
1. TP while ignoring model uncertainty, and
adding only the noise at each time step ($\text{TP}-\sigma_f$). 2. TP
with increased prediction noise 
($\text{TP}+\sigma_n$). We used 300 particles in all cases.

\begin{figure}
	\centering
	\begin{subfigure}{.49\columnwidth}
		\includegraphics[width=1\textwidth]{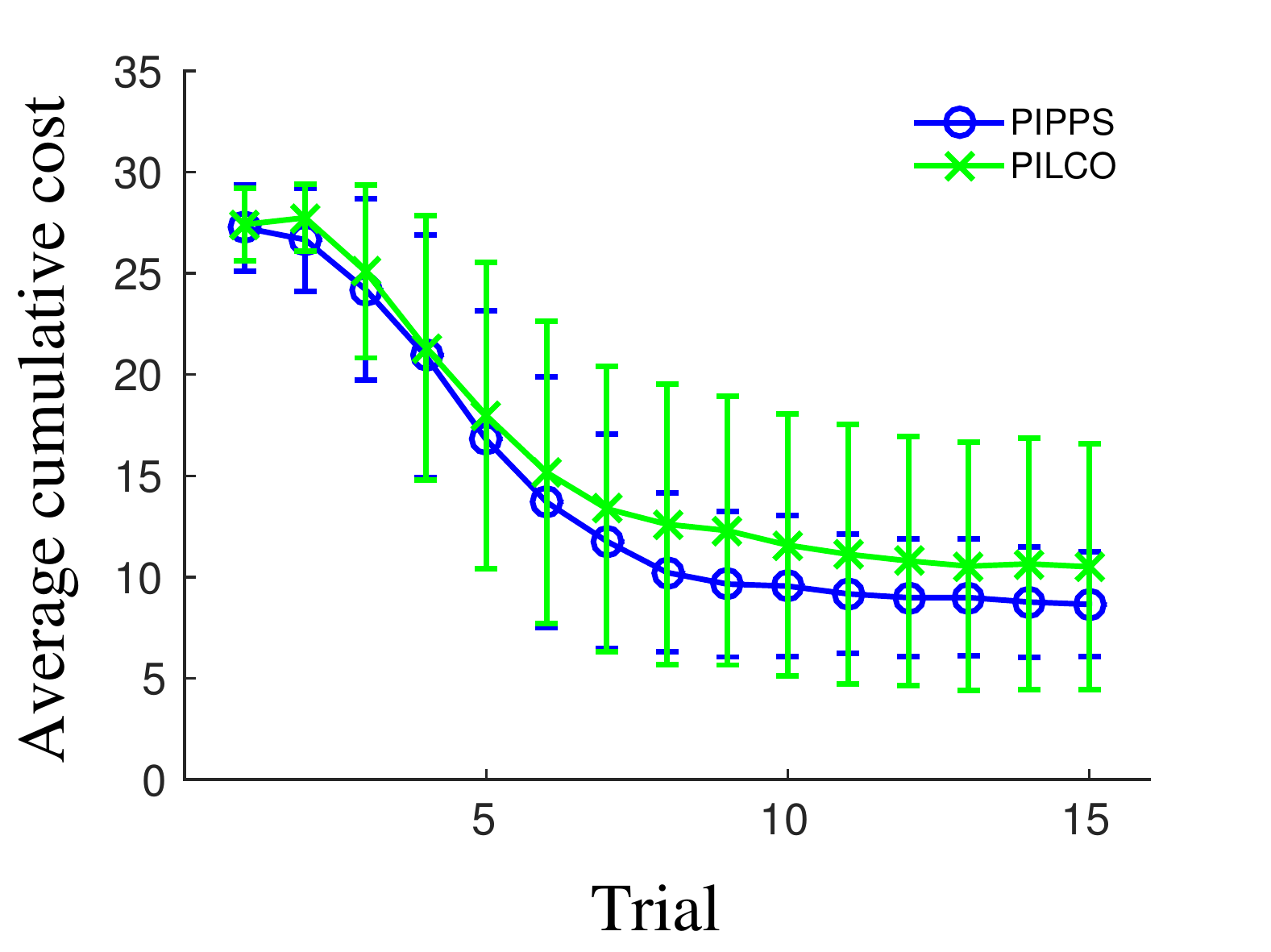}
		\caption{Cart-pole}
	\end{subfigure}
	\begin{subfigure}{.49\columnwidth}
          \includegraphics[width=1\textwidth]{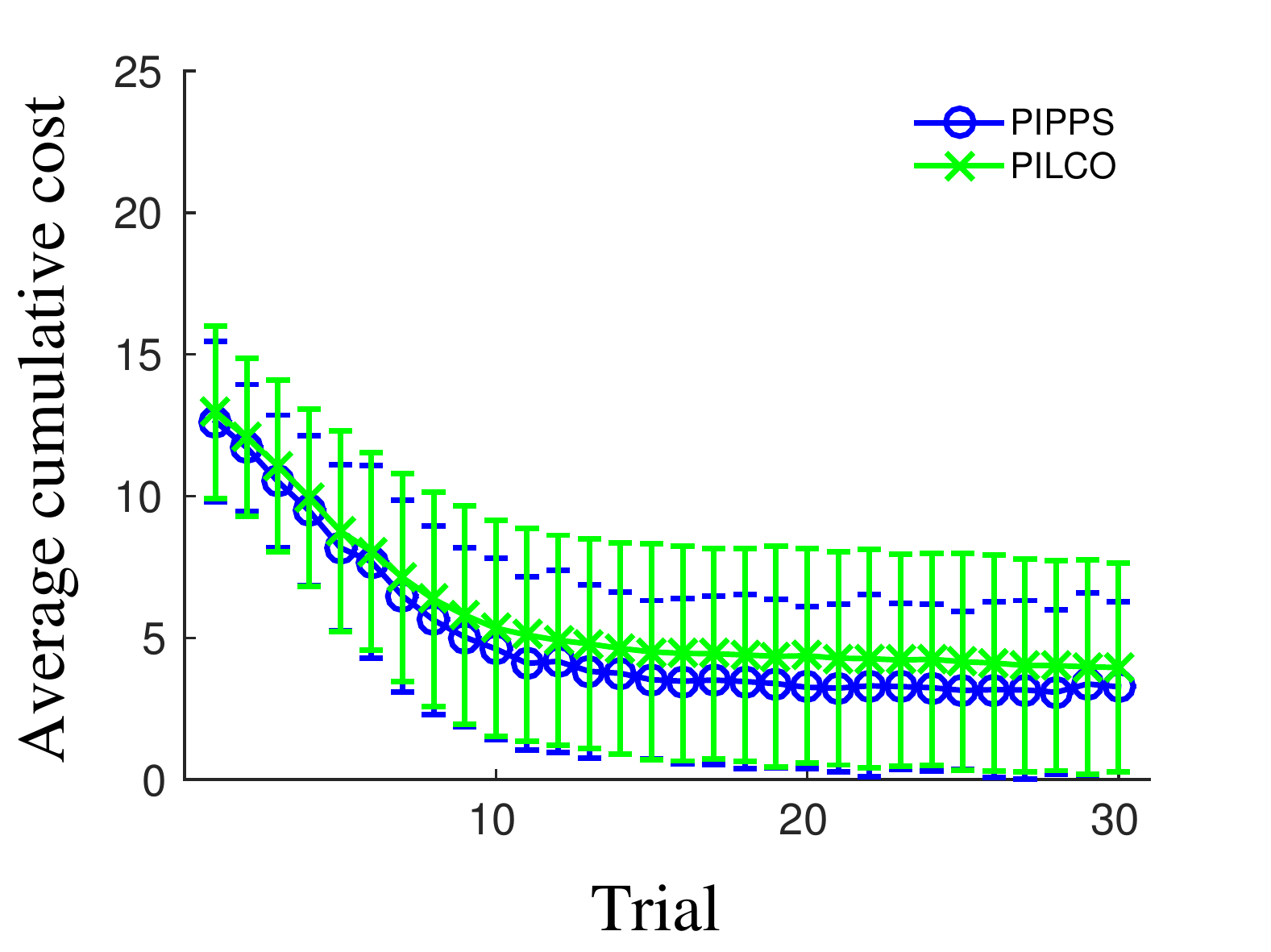}
		\caption{Unicycle}
        \end{subfigure}
        \caption{PIPPS using TP matches PILCO in data-efficiency.}
        \label{expcomparison}
\end{figure}

We performed learning tasks from a recent PILCO paper
\cite{deisenroth2015gppilco}: cart-pole swing-up and balancing, and
unicycle balancing. The simulation dynamics were set to be the same, and
other aspects were similar to the original PILCO.
 The results are in Tables~\ref{cpexps}
and \ref{uniexps}, and in Figure~\ref{expcomparison}.

{\bfseries Common properties in the tasks:} The optimizer was run for 600
policy evaluations between each trial. The SGD 
learning rate, and momentum parameters were $\alpha = 5\times10^{-4}$ and
$\gamma = 0.9$. The episode lengths were
3s for the cart-pole, and 2s for the unicycle. Note that for
the unicycle task, 2s was not sufficient 
for the policy to generalize to long trials, but it still
allowed comparing to PILCO. The control frequencies were 10Hz. The
costs were of the type
$1-\exp(-(\bfv{x}-\bfv{t})^TQ(\bfv{x} - \bfv{t}))$, where $\bfv{t}$ is
the target. The outputs from the policies $\pi(\bfv{x})$ were
constrained by a saturation function:
$\text{sat}(\bfv{u}) = 9\sin(\bfv{u})/8 + \sin(3\bfv{u})/8$, where
$\bfv{u} = \tilde\pi(\bfv{x})$. One experiment
consisted of (1; 5) random trials followed by (15; 30) learned trials for the cart and unicycle tasks respectively. Each experiment was repeated
100 times and averaged. 
Each trial was evaluated by running the policy 30 times, and
averaging, though note that this was performed only for evaluation
purposes -- the algorithms only had access to 1 trial. Success was determined by whether the return of the final trial passed
below a threshold.

\subsubsection{Task Descriptions}

{\bfseries Cart-pole Swing-up and Balancing:} This is a standard
control theory benchmark problem. The task consists of pushing a cart back and forth, to
swing an attached pendulum to an upright position, then  keep it balanced. The state space was
represented as $\bfv{x} = [s, \beta, \dot{s}, \dot{\beta}]$, where $s$
is the cart-position and $\beta$ the pole angle. The base noise
levels were 
$\sigma_{s} = 0.01~\text{m}, \sigma_{\beta} = 1~\text{deg},
\sigma_{\dot{s}} = 0.1~\text{m/s}, \sigma_{\dot{\beta}} =
10~\text{deg/s}$. The noise was modified in different experiments
by a multiplier $k$: $\sigma^2 =
k\sigma^2_{base}$. The original paper considered
direct access to the true state. We set $k = 10^{-2}$ to obtain a
similar setting, but also tested
$k \in \{1,4,9,16\}$. The policy $\tilde\pi$ was a
radial basis function network (a sum of Gaussians) with 50 basis
functions. We considered two cost
functions. One was the same as in the original PILCO,
with $\bfv{x}$ including the sine and cosine, and depended on the distance between the tip of the pendulum to
the position of the tip when the pendulum is
balanced ({\it Tip
  Cost}). The other cost used the raw angle, and had
$Q = \text{diag}([1,1,0,0])$ ({\it Angle Cost}). This cost differs conceptually
from the {\it
  Tip Cost}, because there is only one correct direction in which to
swing up the pendulum.

\begin{table}[t]
\caption{Success rate of learning unicycle balancing}
\vskip 0.1in
\label{uniexps}
\begin{center}
\begin{small}
\begin{sc}
\begin{tabular}{lllllll}%{lcccccccccc}
\toprule
  PILCO & RP & RP$_{\text{FS}}$ & GR & GR$_{\text{FS}}$ & LR & TP\\
\midrule
  {\bf 0.91} & 0.80 & 0.39 & {\bf 0.96} & 0.63 & 0.47 & {\bf 0.94}\\
\bottomrule
\end{tabular}
\end{sc}
\end{small}
\vskip -0.1in
\end{center}
\end{table}

%\captionsetup[table]{justification=raggedright}

\begin{table*}[t]
\caption{Success rate of learning cart-pole swing-up}
\vskip 0.1in
\label{cpexps}
\begin{center}
\begin{small}
\begin{sc}
\begin{tabular}{lllllllllll}%{lcccccccccc}
\toprule
  Cost Function & Noise Multiplier & PILCO & RP & RP$_{\text{FS}}$ & GR & GR$_{\text{FS}}$ & LR & TP &
                      TP$-\sigma_f$ & TP$+\sigma_n$\\
\midrule
Angle Cost & $k = 10^{-2}$ & {\bf 0.88} & 0.69 & 0.24 & 0.63 & 0.74 & 0.57 & {\bf 0.82} & & {\bf 0.96} \\
Angle Cost & $k = 1$ & 0.79 & 0.74 & 0.23 & 0.89 & 0.71 & {\bf 0.96} & {\bf 0.99} & {\bf 0.93} & \\
Angle Cost & $k = 4$ & 0.70 & 0.58 & 0.08 & 0.62 & 0.41 & {\bf 0.94} & {\bf 0.95} & & \\
Angle Cost & $k = 9$ & 0.37 & 0.44 & 0.04 & 0.34 & 0.25 & {\bf 0.86} & {\bf 0.83} & & \\
Angle Cost & $k = 16$ & 0.01 & 0.11 & 0.00 & 0.08 & 0.02 & {\bf 0.45} & {\bf 0.40} & & \\
Tip Cost & $k = 10^{-2}$ & {\bf 0.92} & 0.44 & 0.20 & 0.47 & 0.78 & 0.36 & 0.54 & & \\
Tip Cost & $k = 1$ & {\bf 0.73} & 0.15 & 0.08 & {\bf 0.68} & 0.50 & 0.28 & 0.48 & & \\
\bottomrule
\end{tabular}
\end{sc}
\end{small}
\vskip -0.1in
\end{center}
\end{table*}

{\bfseries Unicycle balancing:} See \cite{deisenroth2015gppilco} for a
description. The task consists of balancing a unicycle robot, with state dimension $D=12$, and control dimension $F=2$. The noise
was set to a low value. The controller $\tilde\pi$ is linear. 

\section{Discussion}
\label{discussion}

\subsection{Learning Experiments}

PILCO performs well in scenarios with no noise, but with noise added
the results deteriorate. This deterioration is most likely caused by
 an accumulation
of errors in the MM approximations, previously observed by
\citet{vinogradska2016stability}, who used quadrature for
predictions. Particles do not suffer from this issue, and using TP
gradients consistently outperforms PILCO with high noise.

On the other hand, at low noise levels, the performance of TP as well
as LR reduces. If all of the particles are sampled from a small
region, it becomes difficult to estimate the gradient from changes in
the return -- in the limit of a delta distribution an LR gradient
could not even be evaluated. The TP gradient is less susceptible to
this problem, because it incorporates information from RP. Finally, if
the uncertainty in predictions is very low (as in $k=10^{-2}$),
one can consider model noise as a parameter that affects learning, and
increase it to acquire more accurate gradients: see
$\text{TP}+\sigma_n$, where the model noise variance was multiplied
by 100.

Notably, approaches which use MM, such as PILCO, and GR outperform
the others when using the {\it Tip Cost}. 
The reason may be the multi-modality of the objective -- with
the {\it Tip Cost}, the pendulum may be swung up from either
direction to solve the task; with the {\it Angle Cost}
there is only one correct direction. Performing MM forces the algorithm along a unimodal path, whereas the particle approach could attempt a bimodal swing-up where some particles go from one side, and
the rest from the other side. Thus, MM
may be performing a kind of "distributional
reward shaping", simplifying the optimization problem. Such an
explanation was previously provided by \citet{gal2016nnpilco}.

Finally, we point to the surprising $\text{TP}-\sigma_f$
experiment. Even though the predictions ignore model uncertainty, the
method achieves 93\% success rate. It is difficult to explain why
learning still worked, but we hypothesize that the success may be
related to the 0 prior mean of the GP. In regions where there is no
data, the mean of the GP dynamics model goes to 0, meaning that the
input control signal has no effect on the particle. Therefore, for the
policy optimization to be successful, the particles would have to be
controlled to stay in regions where there exists data. Note that a
similar result was found by \citet{chatzilygeroudis2017blackdrops} who
used an evolutionary algorithm and achieved 85-90\% success rate at
the cart-pole task even when ignoring model uncertainty.

\subsection{The Curse of Chaos in Deep Learning}

Most machine learning problems involve optimizing the expectation of
an objective function $J(\bfv{x};\theta)$ over some data generating
distribution $p_{Data}(\bfv{x})$, where this distribution can only be
accessed through sample data points $\{\bfv{x}_i\}$.  Our predictive
framework is analogous to a deep model: $p(\bfv{x}_0)$ is the data
generating distribution, $p(\bfv{x}_t;\theta)$ are obtained by pushing
$p_{Data}(\bfv{x})$ through the model layers.  The most common method
of optimization is SGD with pathwise derivatives computed by backpropagation.
Our results suggest that in some situations -- particularly
with very deep or recurrent models -- this approach could degenerate
into a random walk due to an exploding gradient variance.

Exploding gradients have been observed in deep learning research for a
long time \cite{doya1993bifurcations,bengio1994learning}. Typically
this phenomenon is regarded as a numerical issue, which leads to large
steps and unstable learning.  Common countermeasures include gradient
clipping, ReLU activation functions \cite{nair2010ReLU} and smart
initializations. Our explanation to the problem
is different: it is not just that the gradient becomes large, the gradient variance explodes,
meaning that any 
sample from $\bfv{x}_i \sim p_{Data}$ provides essentially no
information about how to change the model parameters $\theta$ to
increase the expectation of the objective over the whole distribution
$\expectw{p_{Data}}{J(\bfv{x})}$.  
While choosing a good initialization is an approach to tackle the
problem, it appears difficult to guarantee that the system does not
become chaotic during learning. For example, in econometrics there are
even cases where the optimal policy may lead to chaotic dynamics
\cite{deneckere1986compchaos}.
Gradient clipping can stop large
parameter steps, but it will not fundamentally solve the problem
if the gradients become random.
Considering that chaos does not
occur in linear systems \cite{alligood1996chaos}, our analysis
suggests a reason for why piece-wise linear activations, which may be
less susceptible to chaos, such as ReLUs perform well in deep
learning.

While we have yet to computationally confirm our deep
hypothesis, several works have
investigated chaos in neural networks
\cite{kolen1991bpchaos,sompolinsky1988chaos},
although we believe we are the first to suggest
that chaos may cause gradients to
degenerate when computed using backpropagation. Notably,
\citet{poole2016expexp} suggested that such properties lead to
"exponential expressivity", but we believe that this phenomenon may instead be a
curse.

\section{Conclusions \& Future Work}
\label{conclusions}

We may have described a limitation of optimizing expectations using
pathwise derivatives, such as those computed by
backpropagation. Moreover, we show a method to counteract this curse
by injecting noise into computations, and using the likelihood
ratio trick. Our {\it total
  propagation algorithm} provides an efficient method to combine
reparameterization gradients on arbitrary stochastic computation
graphs with any amount of other gradient estimators -- even gradients
computed using a value function could be used.
There are countless ways to expand our work: better optimization, incorporate natural
gradients, etc. The flexible nature of our method should make it
easy to extend.

\section*{Acknowledgements}

We thank Chris Reinke and the anonymous reviewers for comments about
clarity. This work was supported by JSPS KAKENHI Grant Number
JP16H06563 and JP16K21738.

\bibliography{part}
\bibliographystyle{icml2018}

\end{document}